\documentclass[a4paper,11pt]{article}

\usepackage{geometry}
\usepackage[utf8]{inputenc}
\usepackage{amsmath,amssymb,amsfonts}
\usepackage{graphicx}
\usepackage{booktabs}
\usepackage[colorlinks=true,linkcolor=blue,citecolor=blue,urlcolor=blue]{hyperref}
\usepackage{caption}
\usepackage{subcaption}
\usepackage{float}
\usepackage{placeins}
\usepackage{authblk}

\usepackage[table]{xcolor}
\usepackage{multirow}
\usepackage{makecell}
\usepackage{siunitx}
\usepackage{pgfplots}
\pgfplotsset{compat=1.18}
\usepackage{algorithm}
\usepackage{algorithmic}
\usepackage{bookmark}
\usepackage{mathptmx}

\title{Med-RADIO: Reducing All Medical Domains Into One via Multi-Teacher Distillation}

\author[1]{Chu Zhang}
\author[1]{Haoyu Jiang}
\author[1]{Hongyuan Zhang}
\author[1]{Hongbin Liu}
\author[1]{Dong Yi}

\affil[1]{Center for Artificial Intelligence and Robotics, Hong Kong Institute of Science and Innovation, Chinese Academy of Sciences, Hong Kong, China}

\date{}

\begin{document}

\maketitle

\begin{abstract}
The rapid expansion of large-scale medical datasets and computational resources has driven significant progress in medical foundation models. Given the inherent heterogeneity of medical imaging modalities, current research mainly follows two paths: specialized models optimized for specific modalities, and generalist models designed to handle multiple modalities. However, medical generalist models suffer from both insufficient training data
scale relative to natural image generalists and inadequate domain-specific depth
relative to medical specialists. Empirically, generalist models establish a cross-modality performance baseline, while specialists define the performance ceiling within their respective domains.
To elevate this baseline toward these ceilings, we propose \textbf{Med-RADIO}, a medical multi-teacher distillation framework that \emph{Reduces All Domains Into One} by compressing complementary expertise from multiple domain-specific teachers into a unified medical vision foundation model. Our method curates both generalist and specialist teachers, allocates modality-aligned distillation streams to reorganize generalist pretraining data so it matches specialist domains, and uses a balanced loss to prevent any single teacher from dominating the distillation process.
On internal and external classification benchmarks spanning five modalities, Med-RADIO improves over strong medical generalists under linear probing and remains competitive with representative specialists on most evaluated modalities.
Code is available at \url{https://github.com/CAIR-HKISI/Med-RADIO}.
\end{abstract}


\begin{figure}[t]
    \centering
    \includegraphics[width=1.0\linewidth]{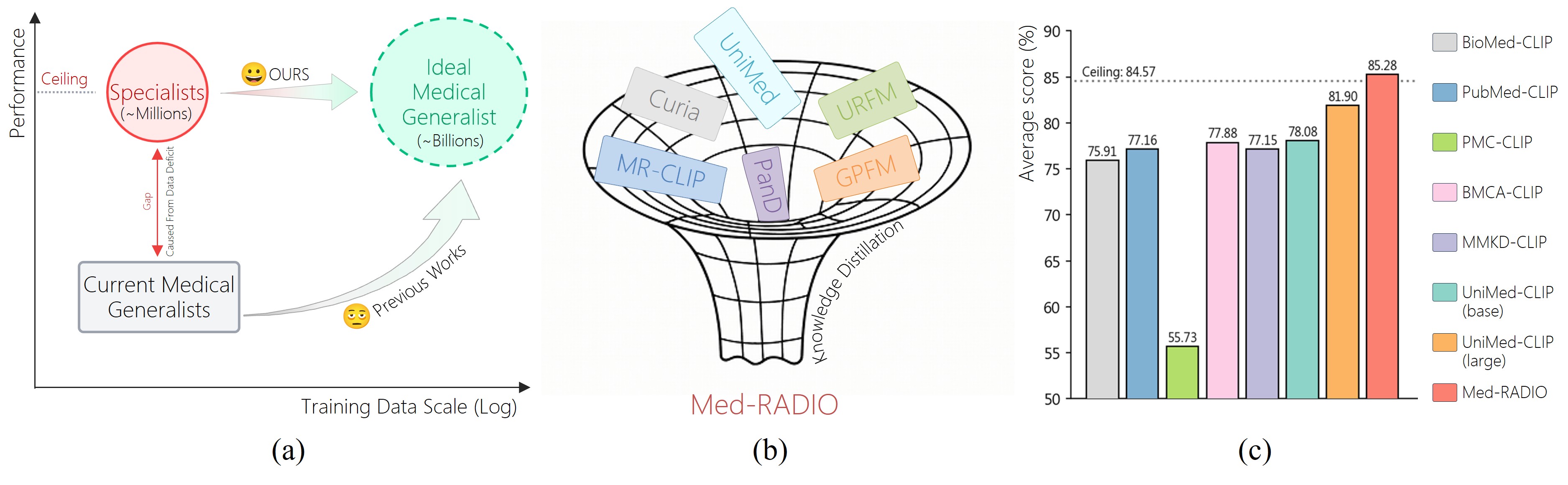}
    \caption{
        Motivation and overview.
        (a)~Generalist--specialist performance gap under limited medical data scale.
        (b)~Med-RADIO compresses heterogeneous teachers into one student encoder.
        (c)~Linear-probing average versus medical generalist baselines (85.28\% for Med-RADIO).
    }
    \label{fig:overview}
\end{figure}

\section{Introduction}
\label{sec:intro}

The rapid expansion of large-scale medical datasets and computational resources has driven significant progress in medical foundation models~\cite{rajendran2025foundation}. Given the inherent heterogeneity of medical imaging modalities, current research follows two trajectories: specialized models optimized for specific modalities such as chest X-ray, pathology, and dermatology, and generalist models designed to handle multiple modalities. However, medical generalist models face a structural weakness rooted in data scale. Compared to well-established generalists in the natural image domain such as CLIP~\cite{radford2021learning,ilharco2021openclip}, DINOv2~\cite{oquab2023dinov2,simeoni2025dinov3}, and SAM~\cite{kirillov2023segment,ravi2024sam,carion2025sam}, they are trained on orders-of-magnitude less data (billions versus millions of images). Consequently, they fail to develop the domain-specific depth comparable to medical specialists trained on millions of modality-specific images. This data deficit restricts generalist models to a cross-modality performance baseline, while specialists define the performance ceiling within their respective domains.

This discrepancy raises a fundamental question: \textit{can we construct a unified medical vision foundation model that elevates generalist capability toward specialist performance?} Several straightforward approaches prove inadequate. Ensembling~\cite{dong2020survey} specialist models incurs prohibitive inference costs and yields no unified representation. Weight merging~\cite{ruan2025task} is infeasible due to architectural heterogeneity across backbones. Multi-teacher knowledge distillation~\cite{you2017learning} offers a principled alternative---training a single student to integrate complementary expertise---yet applying it to medical imaging reveals three critical barriers. First, architectural heterogeneity across medical teachers produces misaligned output spaces. Second, modality heterogeneity introduces distribution sensitivity unique to healthcare: specialist teachers require modality-specific inputs to transfer domain expertise effectively. Third, optimization dynamics tend to be dominated by stronger teachers, suppressing weaker but complementary signals and causing imbalanced capability inheritance~\cite{ranzinger2026c}.

To address these challenges, we propose \textbf{Med-RADIO},  a medical multi-teacher distillation framework that \emph{Reduces All Domains Into One} by compressing complementary expertise from multiple domain-specific teachers into a unified medical vision foundation model. Our key insight is data leverage: by repartitioning the generalist pretraining corpus into modality-specific streams, we indirectly access the distributed expertise of specialists---collectively exposed to tens of millions of domain-specific images---without retrieving original specialist corpora. By freezing all teachers and training only the student encoder with lightweight projectors, we decouple data scale from computational cost, achieving specialist-approaching depth at generalist-scale expense.

Our methodology rests on three innovations. (1) Teacher curation with modality-specific projectors: we systematically select complementary teachers and introduce domain-specific MLP projectors to align heterogeneous output spaces. (2) Modality-aligned distillation streams: we reorganize the pretraining corpus of UniMed-CLIP~\cite{khattak2024unimed} into single-modality subsets, ensuring each specialist receives data drawn from its native distribution. (3) Balanced distillation loss: we employ a contribution-aware weighting scheme that prevents dominant teachers from suppressing complementary signals; UniMed-CLIP further serves as a mandatory generalist anchor, preserving cross-modal stability.

We validate Med-RADIO through extensive experiments on internal and external classification benchmarks spanning five medical imaging modalities. Our unified encoder consistently improves over strong generalist baselines and remains competitive with dedicated specialists on most modalities, demonstrating that strategic distillation enables unified breadth and depth without proportional resource investment. By elevating generalist capability toward specialist performance, Med-RADIO offers a practical path for deploying single, versatile models across heterogeneous medical imaging workflows.

\section{Related Work}

\subsection{Medical Vision Foundation Models}

Large-scale medical vision foundation models have proliferated across radiology, pathology, and multimodal imaging. Early efforts concentrated on modality-specific pretraining, yielding specialized architectures for ultrasound~\cite{kang2025urfm,jiao2024usfm,zhang2025fully}, MRI~\cite{dong2025mri,avci2025mr}, CT~\cite{pai2025vision,hamamci2024developing,dancette2025curia}, chest X-ray~\cite{perez2025exploring,yang2025chest}, fundus photography~\cite{zhou2023foundation,shi2024eyefound,shi2024eyeclip,qiu2023visionfm}, histopathology~\cite{chen2024towards,lu2024visual,ikezogwo2023quilt}, and dermatology~\cite{yan2025multimodal,kim2024transparent}. These specialist models leverage modality-tailored architectures and inductive biases to establish strong performance ceilings within their respective domains.

Concurrently, generalist medical vision models aim to acquire cross-modality representations from large-scale heterogeneous corpora, including vision-language alignment models such as BioMedCLIP~\cite{zhang2023biomedclip}, PubMedCLIP~\cite{eslami2023pubmedclip}, PMC-CLIP~\cite{lin2023pmc}, BMCA-CLIP~\cite{lozano2025biomedica}, and UniMed-CLIP~\cite{khattak2024unimed}, as well as unified multi-modality encoders such as Uni-Med~\cite{zhu2024uni}, Lingshu~\cite{xu2025lingshu}, and Hulu-Med~\cite{jiang2025hulu}. However, these models confront a fundamental structural deficit: relative to natural image foundation models trained on billions of images, medical generalists rely on corpora typically comprising only millions of images. This data gap causes generalist models to consistently underperform modality-specific experts on single-modality benchmarks.

Recent attempts to close this gap via model ensemble~\cite{luo2025ensemble} or weight merging~\cite{yang2025medsamix} encounter fundamental practical barriers: ensemble methods incur multiplicative inference costs incompatible with clinical workflows, while architectural heterogeneity across backbones precludes direct parameter fusion. These limitations motivate distillation-based approaches as a principled alternative.

\subsection{Distilling Foundation Models}

Knowledge distillation~\cite{you2017learning} has recently evolved from model compression toward distillation among foundation models themselves. AM-RADIO~\cite{ranzinger2024radio} and RADIOv2.5~\cite{heinrich2025radiov2} pioneered multi-teacher distillation in natural images, unifying heterogeneous encoders including SAM, CLIP, and DINOv2 into a single versatile backbone. However, these works operate on homogeneous photographic data, circumventing the modality divergence and optimization imbalance intrinsic to medical imaging.

In the medical domain, distillation remains limited in scope: existing efforts distill from single teachers~\cite{patil2025efficient,fu2025sam}, operate within identical architectures~\cite{wang2025unifying}, or restrict transfer to a single modality~\cite{ma2025generalizable}. Med-RADIO instead targets multi-teacher distillation across heterogeneous medical foundation models, jointly addressing architectural misalignment, optimization imbalance, and modality distribution sensitivity.

\subsection{Multi-Teacher Knowledge Distillation}

Multi-teacher knowledge distillation (MKD) aggregates complementary supervisory signals from diverse experts~\cite{you2017learning}. Representation space alignment is critical: directly enforcing feature consistency between structurally dissimilar teachers and students imposes overly rigid constraints~\cite{liu2023simple}. Early uniform weighting schemes treat all teachers equally, risking performance collapse or teacher dominance. Adaptive mechanisms have been proposed to mitigate these issues, including uncertainty-aware weighting~\cite{kendall2018multi}, auxiliary loss-based aggregation~\cite{hu2019learning}, and gradient-balanced adaptive weights~\cite{liu2020adaptive}.

Medical imaging exacerbates these challenges: diverse backbones (ViT variants, Swin Transformers) produce severely misaligned output spaces; heterogeneous pretraining scales cause dominant specialists to suppress complementary generalists; and modality distribution sensitivity requires preserving specialized inductive biases under native data exposure. Med-RADIO addresses these issues with modality-specific projection heads, a contribution-aware balanced loss, and modality-aligned data streams.

\section{Methodology}

\begin{figure}[t]
    \centering
    \includegraphics[width=1.0\linewidth]{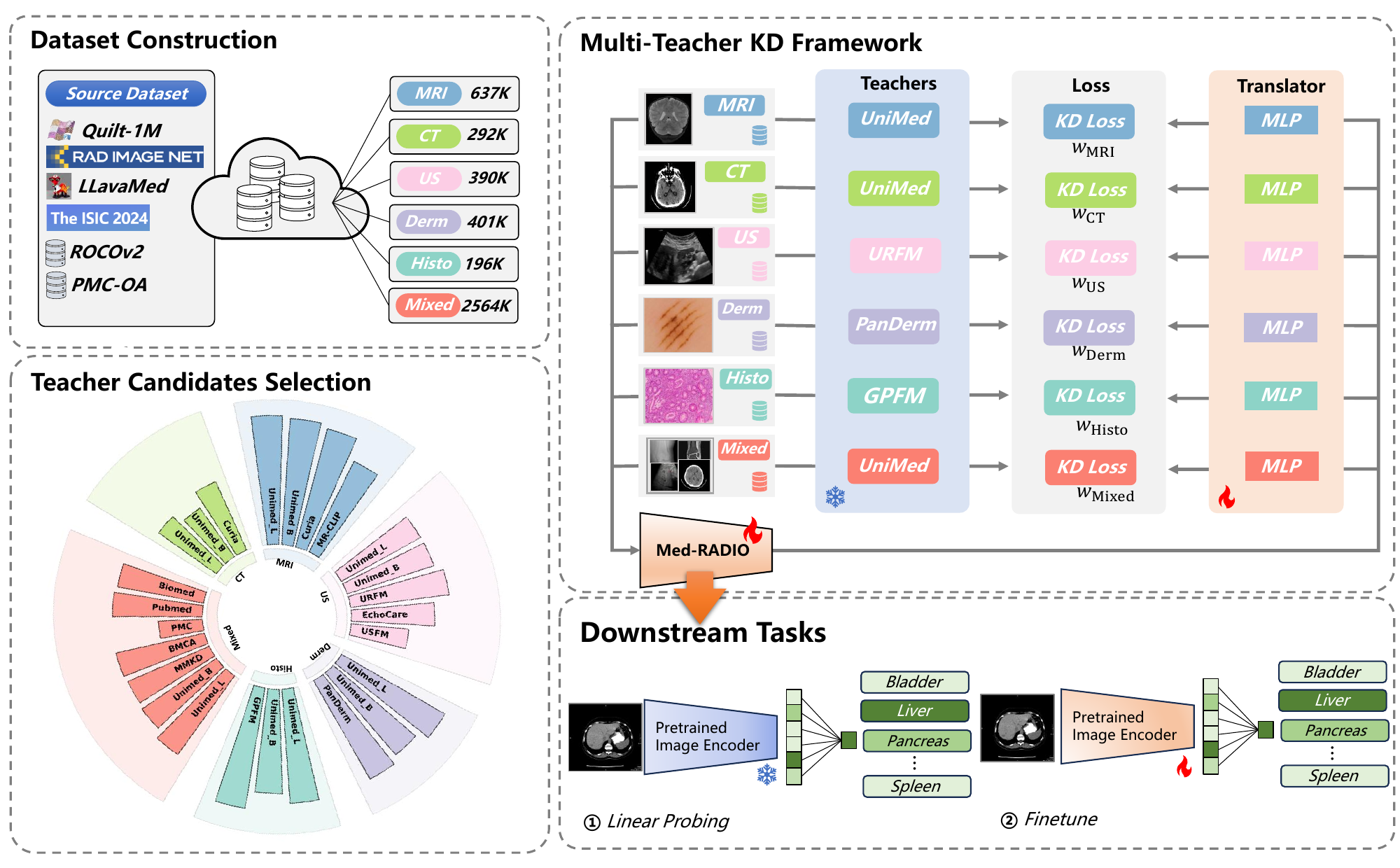}
    \caption{
        Med-RADIO training pipeline: modality-aligned streams, frozen generalist/specialist teachers,
        translator heads, and a dispersion-balanced distillation objective feeding a shared student encoder.
    }
    \label{fig:framework_overview}
\end{figure}

\subsection{Framework Overview}
\label{subsec:framework_overview}
Our objective is to develop a unified medical vision foundation model that synergistically integrates cross-modality generalization capabilities from a generalist teacher with fine-grained, modality-specific expertise from multiple specialist teachers. To achieve this, we propose a multi-teacher distillation framework comprising three interconnected components: (1) a heterogeneous ensemble of pre-trained teacher models, (2) modality-specific translator heads for cross-architecture feature alignment, and (3) a balanced distillation objective. 

As illustrated in Fig.~\ref{fig:framework_overview}, the training pipeline operates through multiple concurrent data streams: each specialist teacher supervises samples from its dedicated modality, while the generalist teacher processes the entire heterogeneous corpus to provide cross-domain regularization. The unified student encoder, shared across all streams, aggregates complementary knowledge through the balanced distillation objective (Eq.~\ref{eq:angular_loss}).
This design lets the student inherit both broad transferable representations and precise domain-specific inductive biases.

\subsection{Pretraining Dataset Construction}
\label{subsec:dataset_construction}
\paragraph{Data Leverage.}
A naive approach to multi-teacher distillation would require curating separate, large-scale pretraining datasets for each specialist teacher, a prohibitively expensive endeavor given the scarcity of annotated medical data. We circumvent this bottleneck through a modality-decoupled data reconstruction strategy: rather than assembling new datasets, we repartition the existing pretraining corpus of UniMed-CLIP into exclusive single-modality subsets, each serving as a dedicated distillation stream for the corresponding specialist teacher. This design enables the generalist teacher to maintain cross-modal alignment on the complete heterogeneous mixture, while each specialist receives purified, modality-focused supervision, without any additional data collection or annotation effort.

\paragraph{Dataset Composition.}
The pretraining data of UniMed-CLIP is composed of subsets encompassing seven major medical imaging modalities: X-ray, computed tomography (CT), magnetic resonance imaging (MRI), ultrasound (US), histopathology whole-slide images (WSI), dermatology photography and Fundus. To support the multi-teacher distillation, we derive five single-modality subsets (US, MRI, CT, histopathology, and dermatology) from this corpus, alongside one multimodal mixture drawn from heterogeneous sources such as PMC-OA\cite{lin2023pmc}, ROCOv2\cite{ruckert2024rocov2}, and LLaVA-Med\cite{li2023llava}. Detailed dataset curation procedures and modality-wise statistics are provided in the Supplementary Material.

\subsection{Teacher Model Composition}
\label{subsec:teacher_composition}

We construct a heterogeneous teacher ensemble $\mathcal{T} = \{T_g\} \cup \{T_m\}_{m \in \mathcal{M}}$ 
comprising one generalist medical vision foundation model and multiple modality-specific specialists. 
Formally, let $\mathcal{M} = \{\mathrm{US}, \mathrm{CT}, \mathrm{MRI}, \mathrm{Path}, \mathrm{Derm}\}$ 
denote the set of specialist-covered modalities. The generalist teacher $T_g$ provides cross-modality 
representations $f_g: \mathcal{X} \rightarrow \mathbb{R}^{D_g}$, while each specialist teacher $T_m$ 
offers modality-specific inductive biases through $f_m: \mathcal{X}_m \rightarrow \mathbb{R}^{D_m}$. 
The student encoder $f_s(\cdot; \theta)$ is trained solely via multi-teacher distillation:

\begin{equation}
\min_{\theta}\ \mathcal{L}(\theta) = \mathcal{L}_{\mathrm{KD}}^{(g)}(\theta) 
+ \sum_{m \in \mathcal{M}} \mathcal{L}_{\mathrm{KD}}^{(m)}(\theta),
\label{eq:total_loss}
\end{equation}
where the per-teacher losses $\mathcal{L}_{\mathrm{KD}}^{(g)}$ and $\{\mathcal{L}_{\mathrm{KD}}^{(m)}\}$ 
are detailed in Sec.~\ref{subsec:loss_formulation}.

\subsection{Modality-Specific Translator Heads}
\label{subsec:translator_heads}

Direct distillation from heterogeneous teachers is impeded by discrepancies in backbone 
architecture, feature dimensionality, and representation semantics. To bridge these gaps 
without modifying the frozen teachers, we introduce lightweight translator heads 
that project student features into each teacher's output space.

For each teacher $m$, we instantiate a dedicated three-layer MLP projector 
$\phi_m: \mathbb{R}^{d_s} \rightarrow \mathbb{R}^{d_m}$:

\begin{equation}
\phi_m(\mathbf{z}) = \mathbf{W}_3 \cdot \sigma\!\left(\mathbf{W}_2 \cdot 
\sigma\!\left(\mathbf{W}_1 \mathbf{z}\right)\right),
\label{eq:translator}
\end{equation}
where $d_s$ is the student's output dimension, $d_m$ is teacher $m$'s output dimension, 
all hidden layers use a fixed width of 2048, and $\sigma$ denotes the SiLU activation. 
The student maintains two sets of projectors: generalist projectors $\{\phi_m^g\}$ 
aligned to the generalist teacher, and specialist projectors $\{\phi_m^s\}$ aligned 
to the modality-specific specialists, each instantiated with its respective target dimension $d_m$.

This design decouples representation alignment from feature extraction: the shared student 
encoder $f_\theta$ learns a unified medical representation, while each $\phi_m$ adapts that 
representation to its teacher's geometry. All translator heads are discarded after pretraining; 
downstream tasks use only the bare encoder $f_\theta$, incurring no additional inference overhead.

\subsection{Balanced Distillation Loss}
\label{subsec:loss_formulation}

A central challenge in multi-teacher distillation is teacher collapse---where a 
dominant teacher suppresses complementary signals. Rather than relying on dynamic weighting 
schemes, we address this through a statistics-based balancing mechanism\cite{ranzinger2026c} built into 
each per-teacher loss $\mathcal{L}_{\mathrm{KD}}^{(m)}$ in Eq.~\ref{eq:total_loss}: each 
loss is normalized by its teacher's feature dispersion, estimated offline from the 
corresponding modality stream prior to training.

% \paragraph{Angular Distillation Loss.}
Rather than minimizing raw feature distances, we supervise the geometric structure
of the feature space via an angular loss. Given student feature $\mathbf{x}$ and teacher 
feature $\mathbf{y}$, the angular distance is:

\begin{equation}
\Theta(\mathbf{x}, \mathbf{y}) = \arccos\!\left(\frac{\mathbf{x}^\top\mathbf{y}}
{\|\mathbf{x}\|\|\mathbf{y}\|}\right).
\label{eq:angular_distance}
\end{equation}

To make this loss scale-invariant across teachers with different feature variances, we 
normalize by the intrinsic angular dispersion of each teacher's distribution. Let 
$\boldsymbol{\mu}_{\mathbf{y}} = {\mathbb{E}[\mathbf{y}]}/{\|\mathbb{E}[\mathbf{y}]\|}$ 
denote the mean direction of teacher features; the angular dispersion is:

\begin{equation}
\mathrm{Disp}(\Theta_{\mathbf{y}}) = \mathbb{E}\!\left[\Theta\!\left(\mathbf{y},\, 
\boldsymbol{\mu}_{\mathbf{y}}\right)^2\right],
\label{eq:angular_dispersion}
\end{equation}
estimated offline on $10{,}000$ samples from each teacher's modality stream 
(Tab.~\ref{tab:angular_disp} in the Supplementary Material). The normalized angular loss is:

\begin{equation}
\mathcal{L}_{\mathrm{KD}}^{(m)}(\mathbf{x}, \mathbf{y}) = \frac{\Theta(\mathbf{x}, 
\mathbf{y})^2}{\mathrm{Disp}(\Theta_{\mathbf{y}})}.
\label{eq:angular_loss}
\end{equation}

This static normalization simultaneously amplifies suppressed low-dispersion signals and 
attenuates over-dominant high-dispersion ones, achieving equitable gradient magnitudes 
across teachers without manual weight tuning or online adaptation.

\section{Experiments}

\subsection{Implementation Details}
\label{sec:implementation_details}

\paragraph{Teacher Selection.}
We select the specialist teacher for each modality by evaluating candidate modality-specific models under linear probing and choosing the top performer, while excluding UniMed-CLIP variants from specialist selection because they already serve as the mandatory generalist anchor. Detailed scores are provided in the Supplementary Material (Tab.~\ref{tab:teacher_selection}).

\paragraph{Student Architecture and Training Configuration.}
The student encoder is ViT-Large/16 (1024-d hidden dimension, 24 transformer 
layers, 16 attention heads, $224\times224$ input resolution), initialized with standard ImageNet-21k pretraining~\cite{dosovitskiy2020image}. Each modality-specific translator head 
consists of an MLP followed by LayerNorm, mapping the student's 1024-d CLS token to the 
corresponding teacher's output dimensionality; all translator heads are discarded after 
pretraining, and downstream evaluations use solely the bare student backbone.

Training is conducted with PyTorch and Accelerate across 6$\times$A100 (80\,GB) GPUs. 
The student is optimized for 10 epochs using AdamW ($\beta_1{=}0.9$, $\beta_2{=}0.999$, 
weight decay $0.05$) with a cosine annealing schedule (peak LR $1\times10^{-4}$) and a 
1-epoch linear warm-up. The global batch size is 192 (32 per teacher stream), assembled 
via proportional multi-stream sampling where each modality contributes samples according 
to its dataset fraction. Teacher-specific data augmentation is applied independently per stream.

\paragraph{Teacher Feature Statistics and Balancing Rationale.}
As formalized in Eq.~\ref{eq:angular_loss}, each teacher loss is normalized by its offline angular dispersion $\mathrm{Disp}(\Theta_{\mathbf{y}})$ (Eq.~\ref{eq:angular_dispersion}), amplifying low-dispersion teachers and attenuating high-dispersion ones. Estimates on $10{,}000$ samples per stream are reported in Tab.~\ref{tab:angular_disp} in the Supplementary Material.

\paragraph{Downstream Evaluation.}
We assess representational quality under two protocols:
1) linear probing: the pre-trained encoder is strictly frozen; only a linear head trained on top of the extracted global CLS token is optimized for 50 epochs with AdamW (peak learning rate $1\times10^{-3}$, cosine schedule, batch size 256), directly probing the linearly separable structure of the pre-trained representations.
2) full fine-tuning: all encoder parameters are jointly optimized end-to-end with a task-specific head for 50 epochs (peak LR $5\times10^{-5}$, cosine schedule, batch size 64), providing an upper-bound assessment of adaptation capacity.

\paragraph{Benchmarks and Metrics.}
Our primary (\emph{internal}) evaluation spans nine classification benchmarks across five modalities: two ultrasound datasets (Thyroid-US and Breast-US), two MRI datasets (ACL Tear and Meniscus Tear), three CT datasets (MediMeTA Axial, Coronal, and Sagittal), one histopathology dataset (PCam), and one dermatology dataset (HAM). Following domain conventions, we report AUC for ultrasound/MRI and ACC for CT, histopathology, and dermatology.
To assess generalization beyond this internal suite, we additionally evaluate on an \emph{external} five-modality classification protocol adapted from MMKD-CLIP~\cite{wang2025unifying} (Tab.~\ref{tab:internal_external}).
Further details on the internal datasets are provided in the Supplementary Material.

\subsection{Main Results}
\label{sec:main_results}

We benchmark Med-RADIO against two groups of baselines: (i) medical generalist models and (ii) modality-specific specialist models, under both linear probing and full fine-tuning.

\paragraph{Comparison with Medical Generalist Baselines.}
Tab.~\ref{tab:lp_cls} reports linear probing results across the nine classification benchmarks.
\begin{table}[htbp]
\centering
\caption{Linear probing (\%) on nine classification benchmarks. Best in \textbf{bold}. AUC for US/MRI; ACC for CT/Histo./Derm.}
\label{tab:lp_cls}
\resizebox{\linewidth}{!}{%
\begin{tabular}{l|cccccccccc}
\toprule
\multirow{2}{*}{Model} 
  & \multicolumn{2}{c}{US} 
  & \multicolumn{2}{c}{MRI} 
  & \multicolumn{3}{c}{CT} 
  & \multicolumn{1}{c}{Histo} 
  & \multicolumn{1}{c}{Derm} 
  & \multirow{2}{*}{Avg} \\
\cmidrule(lr){2-3}\cmidrule(lr){4-5}\cmidrule(lr){6-8}\cmidrule(lr){9-9}\cmidrule(lr){10-10}
 & Thyroid & Breast & ACL & Meniscus 
 & \makecell{Axial} & \makecell{Coronal} & \makecell{Sagittal} 
 & PCam & \makecell{HAM} & \\
\midrule
Biomed-CLIP \cite{zhang2023biomedclip}          & 78.25 & 81.79 & 89.47 & 86.99 & 67.64 & 57.93 & 54.05 & 83.40 & 83.71 & 75.91 \\
Pubmed-CLIP \cite{eslami2023pubmedclip}          & 78.36 & 84.92 & 90.91 & 88.68 & 72.33 & 59.39 & 54.85 & 82.43 & 82.62 & 77.17 \\
PMC-CLIP \cite{lin2023pmc}             & 56.96 & 66.72 & 68.75 & 74.61 & 37.86 & 34.14 & 34.95 & 75.35 & 52.25 & 55.73 \\
BMCA-CLIP \cite{lozano2025biomedica}            & 82.57 & 82.10 & 93.42 & 92.06 & 69.90 & 61.65 & 51.78 & 85.56 & 81.88 & 77.88 \\
MMKD-CLIP \cite{wang2025unifying}           & 63.63 & 76.52 & 97.30 & 94.91 & 72.01 & 62.30 & 56.63 & 85.69 & 85.37 & 77.15 \\
UniMed-CLIP (base) \cite{khattak2024unimed}   & 76.96 & 79.34 & 97.30 & 95.08 & 67.80 & 59.22 & 55.34 & 85.41 & 86.26 & 78.08 \\
UniMed-CLIP (large) \cite{khattak2024unimed}  & 79.18 & 77.12 & 97.05 & \textbf{97.53} & 76.38 & 65.70 & 62.78 & \textbf{90.23} & 91.15 & 81.90 \\
Med-RADIO (Ours)          & \textbf{84.91} & \textbf{88.72} & \textbf{98.16} & 95.49 & \textbf{80.42} & \textbf{71.20} & \textbf{67.48} & 88.32 & \textbf{92.86} & \textbf{85.28} \\
\bottomrule
\end{tabular}%
}
\end{table}

Under linear probing (Tab.~\ref{tab:lp_cls}), Med-RADIO achieves a mean AUC/ACC of \textbf{85.28\%}, surpassing the strongest generalist UniMed-CLIP (large) by \textbf{+3.38\,pp} and all other baselines by over 7\,pp. Gains are most pronounced in ultrasound and CT, the modalities where specialist priors are most complementary to generalist pretraining, with Breast-US AUC improving by +11.60\,pp (the largest margin) and CT Axial/Coronal/Sagittal ACC by $+$4.0/$+$5.5/$+$4.7\,pp, respectively. The two benchmarks where Med-RADIO falls below UniMed-CLIP (large), namely MRI Meniscus AUC ($-$2.0\,pp) and PCam ACC ($-$1.9\,pp), both involve fine-grained texture discrimination that is difficult to capture through global CLS supervision under the strictly frozen evaluation protocol.

\paragraph{Comparison with Modality-Specific Specialists.} 
\begin{figure}[htbp]
    \centering
    \includegraphics[width=0.9\linewidth]{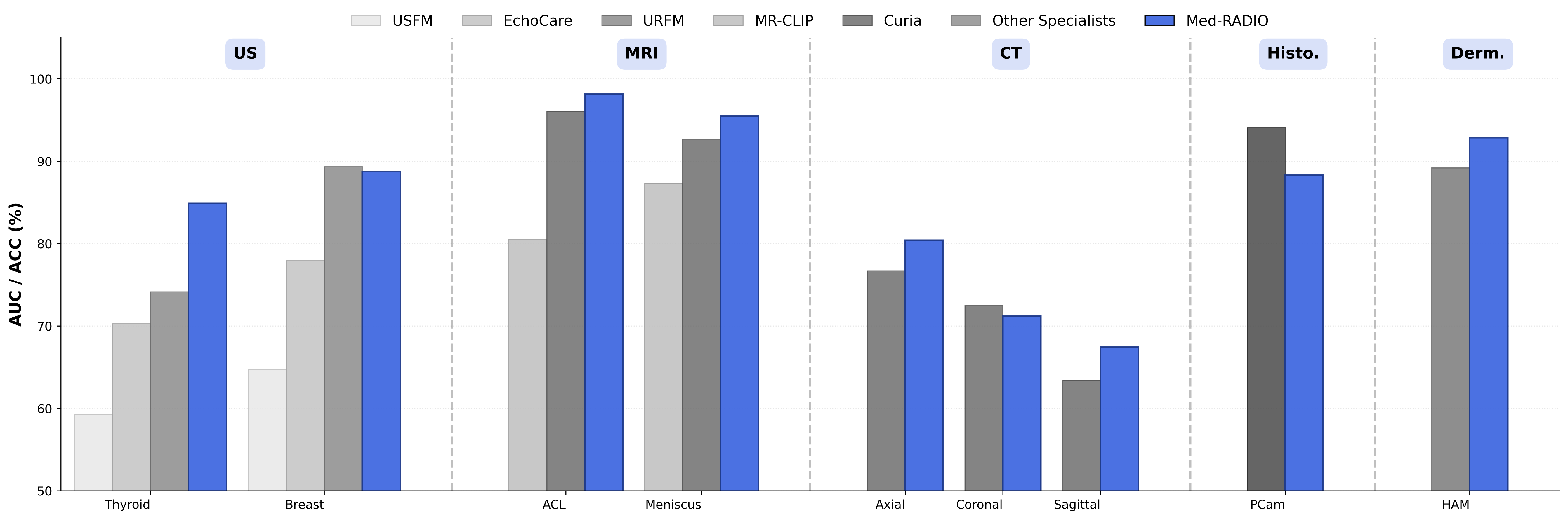}
    \caption{Linear probing versus the strongest dedicated specialist per modality. Bars show absolute AUC/ACC; $\Delta$ above each group is Med-RADIO minus the best specialist.}
    \label{fig:specialist_lp}
\end{figure}
Fig.~\ref{fig:specialist_lp} compares Med-RADIO with dedicated specialists under linear probing.
Med-RADIO matches or exceeds the best specialist on four of five modalities.
The largest gains appear on ultrasound Thyroid ($+10.8$\,pp over URFM) and CT Sagittal/Axial ($+4.1$/$+$3.7\,pp over Curia); MRI and dermatology also favor Med-RADIO, with only a negligible Breast deficit ($-0.6$\,pp).
The main deficit remains histopathology ($-5.8$\,pp vs.\ GPFM), consistent with the mismatch between global CLS distillation and WSI patch-level specialist pretraining.

\begin{figure}[htbp]
    \centering
    \includegraphics[width=0.75\linewidth]{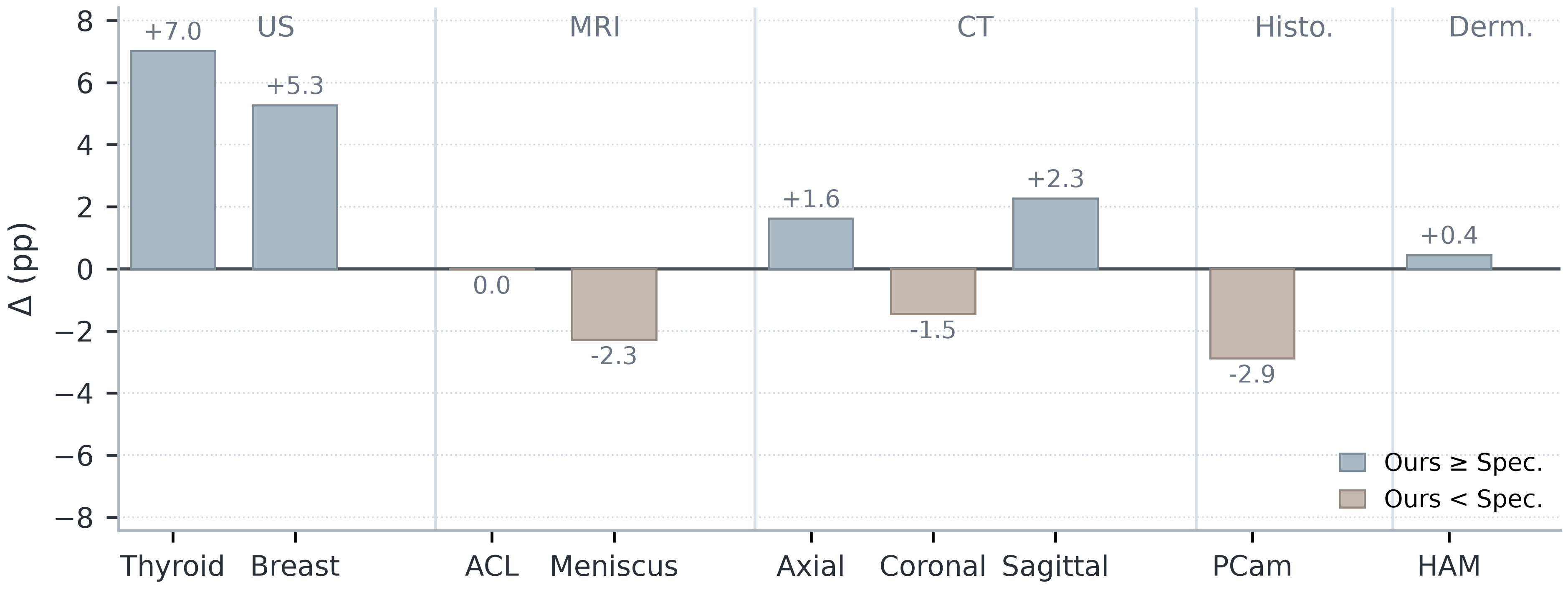}
    \caption{Fine-tuning $\Delta$ (Med-RADIO $-$ best specialist per modality; pp). Positive favors Med-RADIO.}
    \label{fig:specialist_ft}
\end{figure}

Fig.~\ref{fig:specialist_ft} reports the corresponding fine-tuning gaps.
Med-RADIO leads on five of nine tasks, especially ultrasound (Thyroid/Breast $+7.0$/$+$5.3\,pp), with smaller gains on CT Sagittal/Axial and HAM.
The largest deficits remain PCam and Meniscus (about $-2$--$3$\,pp), mirroring the linear-probing histopathology gap.

\paragraph{Comparison with MTKD and Balancing Baselines.}
Tab.~\ref{tab:internal_external} compares Med-RADIO with MMKD-CLIP~\cite{wang2025unifying} and matched balancing variants (Uniform / Uncertainty~\cite{kendall2018multi} / AdaLoss~\cite{hu2019learning}) under the internal/external protocol defined above; per-modality scores average the corresponding benchmarks.
Med-RADIO attains the best mean (\textbf{84.66\%}), beating MMKD-CLIP by \textbf{+3.71\,pp} and the strongest balancing baseline (Uncertainty, 84.14\%) by $+0.52$\,pp.
Gains over MMKD-CLIP concentrate on internal ultrasound/CT, where modality-aligned specialists help most; local trade-offs versus Uncertainty/AdaLoss on a few cells do not overturn the average advantage.

\begin{table}[htbp]
\centering
\caption{Internal/external linear probing (\%). External tasks are held out from teacher-pool construction; balancing variants share the Med-RADIO recipe and differ only in loss weighting. Best Avg in \textbf{bold}.}
\label{tab:internal_external}
\resizebox{\linewidth}{!}{%
\begin{tabular}{ll|ccccc|ccccc|c}
\toprule
% Type & Model & \multicolumn{5}{c|}{Internal} & \multicolumn{5}{c|}{External} & Avg \\
Type & Model & \multicolumn{5}{c|}{Internal} & \multicolumn{5}{c|}{External} & Avg \\
\cmidrule(lr){3-7}\cmidrule(lr){8-12}
 & & US & MRI & CT & Histo. & Derm. & US & MRI & CT & Histo. & Derm. & \\
\midrule
MTKD & MMKD-CLIP~\cite{wang2025unifying}
 & 70.08 & 96.11 & 63.65 & 85.72 & 87.87
 & 78.17 & 99.96 & 99.69 & \textbf{77.00} & 70.02 & 80.95 \\
\midrule
Balancing & Uniform
 & 83.56 & 95.93 & 73.25 & 86.92 & 90.18
 & 79.66 & 99.98 & 98.97 & 71.00 & 75.11 & 82.73 \\
 & Uncertainty~\cite{kendall2018multi}
 & 84.73 & 96.49 & \textbf{74.38} & 88.64 & \textbf{93.76}
 & 81.39 & 100.00 & 99.61 & 73.00 & 77.21 & 84.14 \\
 & AdaLoss~\cite{hu2019learning}
 & 82.89 & 95.87 & 73.30 & 89.15 & 94.21
 & \textbf{82.52} & 99.99 & 99.63 & 75.00 & 77.61 & 83.95 \\
\midrule
Ours & Med-RADIO
 & \textbf{87.11} & \textbf{96.69} & 72.98 & \textbf{89.64} & 93.44
 & 81.04 & \textbf{100.00} & \textbf{99.69} & 76.00 & \textbf{78.25} & \textbf{84.66} \\
\bottomrule
\end{tabular}%
}
\end{table}

\subsection{Ablation Studies and Analysis}
\begin{figure}[htbp]
    \centering
    \includegraphics[width=0.8\linewidth]{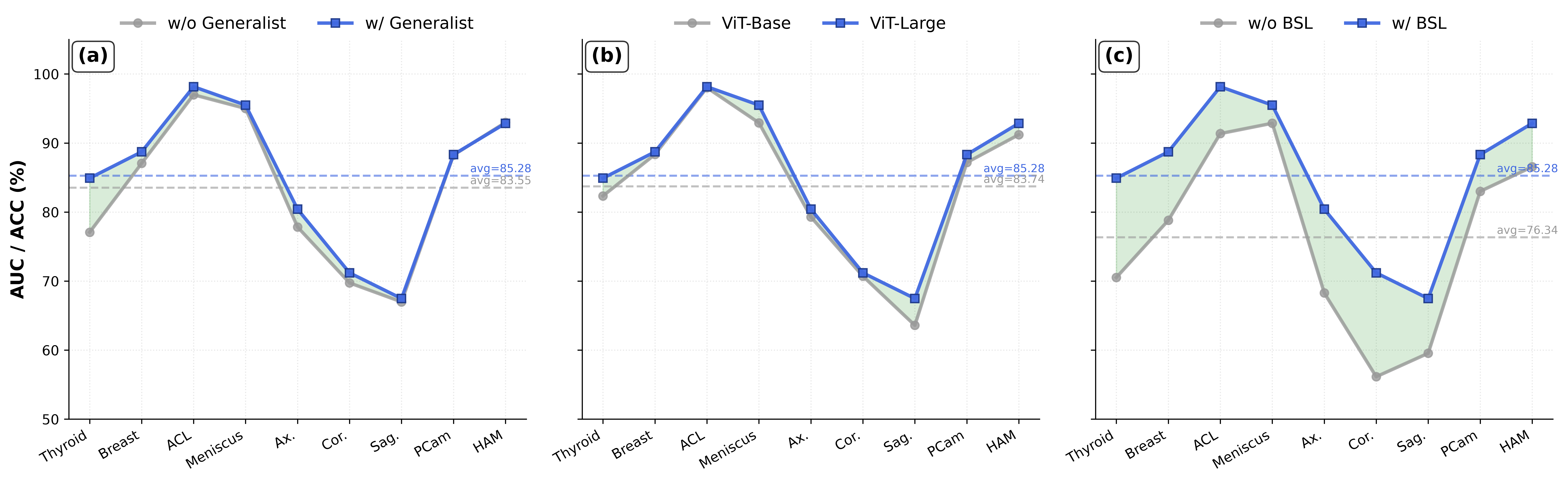}
    \caption{Linear-probing ablations on nine benchmarks.
    Each panel contrasts two settings (shaded regions mark where the second legend entry is higher/lower).
    (a)~w/ vs.\ w/o generalist teacher;
    (b)~ViT-Large vs.\ ViT-Base student;
    (c)~w/ vs.\ w/o dispersion-balanced loss.}
    \label{fig:three_ablations}
\end{figure}

We conduct controlled ablations to quantify the individual contribution of each design component in Med-RADIO. All experiments follow the linear probing protocol on the nine benchmarks, with results shown in Fig.~\ref{fig:three_ablations}.

\paragraph{Effect of the Generalist Teacher (Fig.~\ref{fig:three_ablations}(a)).}
Removing the generalist while keeping all specialists lowers the mean by $-$1.74\,pp (85.28\%$\to$83.55\%), mainly on ultrasound Thyroid ($-$7.84\,pp) and CT Axial.
Histopathology is nearly unchanged, indicating that the generalist primarily supplies cross-modal regularization where specialist coverage is thinner.

\paragraph{Effect of Student Model Scale (Fig.~\ref{fig:three_ablations}(b)).}
Scaling from ViT-Base to ViT-Large yields $+1.55$\,pp on average, with the largest lifts on CT Sagittal and MRI Meniscus, suggesting that higher capacity better absorbs heterogeneous teacher spaces.

\paragraph{Effect of the Balanced Loss Strategy (Fig.~\ref{fig:three_ablations}(c)).}
Dispersion balancing is the most critical ingredient: removing it collapses the mean by $-$8.94\,pp (85.28\%$\to$76.34\%), with severe drops on CT and ultrasound Thyroid.
Without it, high-dispersion teachers (e.g., Curia on CT) dominate and suppress complementary streams, confirming that equitable weighting is required for heterogeneous multi-teacher distillation.

\section{Discussion and Limitation}
Our results support an affirmative answer to the motivating question: a single student encoder can elevate medical generalist baselines toward specialist ceilings without ensembling models or merging incompatible weights.
Under linear probing, Med-RADIO reaches \textbf{85.28\%} mean AUC/ACC (\textbf{+3.38\,pp} over UniMed-CLIP large), remains competitive with dedicated specialists on four of five modalities, and outperforms both a medical MTKD baseline and matched balancing alternatives on the internal/external protocol.
Taken together, these findings indicate that the breadth--depth dilemma is amenable to data leverage: by repartitioning an existing generalist corpus into modality-aligned streams and distilling frozen heterogeneous teachers through lightweight translators, specialist priors can be absorbed at generalist-scale training cost and deployed in a single forward pass.

Several practical takeaways emerge from the ablations and baseline comparisons.
A mandatory generalist teacher is not redundant with specialists: removing it degrades average performance by $-$1.74\,pp and hits ultrasound/CT hardest, consistent with a cross-modal regularizer that stabilizes domains where specialist coverage alone is insufficient.
Contribution-aware balancing is likewise essential: unweighted multi-teacher optimization is vulnerable to dispersion extremes (low-dispersion teachers such as URFM being drowned out; high-dispersion teachers such as Curia/GPFM dominating), and generic adaptive schemes (Uncertainty, AdaLoss) underperform our angular-dispersion weighting under an otherwise identical recipe.
Finally, standard medical MTKD without modality-native teacher assignment (e.g., MMKD-CLIP) lags most clearly on ultrasound and CT, reinforcing that architectural alignment alone does not substitute for distribution-matched supervision.

Nonetheless, important limitations remain.
(i) Histopathology stream quality.
The largest specialist gap appears on histopathology, which we attribute primarily to a data-quality mismatch:
GPFM is pretrained on curated pathology corpora, whereas our histopathology distillation stream comes from Quilt-1M~\cite{ikezogwo2023quilt}, a noisier YouTube-collected collection.
Closing this gap likely requires higher-quality, distribution-matched pathology data (and optionally multi-scale supervision), beyond global CLS distillation alone.
(ii) Coverage.
Our evaluation focuses on 2D classification over five modalities; chest X-ray, fundus, volumetric 3D understanding, dense prediction, and VQA are not yet systematically assessed.
(iii) Teacher- and protocol-dependence.
Student quality remains bounded by the frozen teacher pool and public stream corpora, and our objective emphasizes CLS angular matching under frozen linear probing, which under-emphasizes fine-grained texture cues (e.g., Meniscus, PCam).
External hold-outs mitigate but do not remove this coupling; clinical claims should therefore be read as inference-efficient unification pending broader validation.

\section{Conclusion}
We presented \textbf{Med-RADIO}, a multi-teacher distillation framework that compresses a cross-modal generalist and modality-specific specialists into one vision encoder via translator heads, modality-aligned data streams, and contribution-aware balanced loss.
Trained on a restructured 4.5M-image corpus without new specialist corpora, it achieves \textbf{85.28\%} mean linear-probing AUC/ACC (\textbf{+3.38\,pp} over the strongest generalist) and matches or surpasses dedicated specialists on four of five modalities in a single forward pass.
These results show that principled heterogeneous distillation offers a practical alternative to ensembles and weight merging for unifying medical vision foundations.
Future work will extend Med-RADIO to additional modalities, volumetric inputs, and multi-scale pathology supervision.

\FloatBarrier
\bibliographystyle{splncs04}
\bibliography{main}

\newpage

\section*{Supplementary Material}

\setcounter{table}{0}
\renewcommand{\thetable}{S\arabic{table}}

\subsubsection*{Pretraining Corpus Reconstruction}

Our pretraining corpus follows the modality-decoupled reconstruction principle: we disaggregate UniMed-CLIP's original 5.3M multi-modal mixture into six independent data streams, enabling modality-specific teacher assignment during distillation. Tab.~\ref{tab:supp_pretrain} details the composition. Single-modality streams (US, MRI, CT, Histo., Derm.) are sourced from public domain-specific datasets with expert annotations, while the multimodal mixture stream aggregates cross-domain image-text pairs from PMC-OA~\cite{lin2023pmc}, ROCOv2~\cite{ruckert2024rocov2}, and LLaVA-Med~\cite{li2023llava}. This reconstruction ensures that specialist teachers supervise only their respective modalities, while the generalist teacher provides cross-modal regularization across all streams.

\begin{table}[h]
\centering
\caption{Pretraining stream composition (4.54M pairs). Specialist streams are single-modality; Mix is supervised by the generalist.}
\label{tab:supp_pretrain}
\small
\setlength{\tabcolsep}{4pt}
\begin{tabular}{lrll}
\toprule
\textbf{Stream} & \textbf{\#Pairs} & \textbf{Sources} & \textbf{Teacher} \\
\midrule
US   & 390k  & RadImageNet (US subset)~\cite{mei2022radimagenet} & URFM~\cite{kang2025urfm} \\
MRI  & 673k  & RadImageNet (MRI subset)~\cite{mei2022radimagenet} & Curia~\cite{dancette2025curia} \\
CT   & 290k  & RadImageNet (CT subset)~\cite{mei2022radimagenet} & Curia~\cite{dancette2025curia} \\
Histo. & 196k & Quilt-1M~\cite{ikezogwo2023quilt} & GPFM~\cite{ma2025generalizable} \\
Derm. & 401k  & ISIC2024 archive~\cite{kurtansky2024slice3d} & PanDerm~\cite{yan2025multimodal} \\
\midrule
Mix  & 2.59M & \makecell[l]{PMC-OA~\cite{lin2023pmc}, ROCOv2~\cite{ruckert2024rocov2}, \\ LLaVA-Med~\cite{li2023llava}} & \makecell[l]{UniMed-CLIP~\cite{khattak2024unimed}} \\
\midrule
\textbf{Total} & \textbf{4.54M} & --- & --- \\
\bottomrule
\end{tabular}
\end{table}

\paragraph{Preprocessing Protocol.}
All images are center-cropped to $224{\times}224$ resolution with ImageNet normalization ($\mu{=}[0.485, 0.456, 0.406]$, $\sigma{=}[0.229, 0.224, 0.225]$). Grayscale medical images (ultrasound, MRI, CT) are channel-replicated to RGB format for ViT compatibility. No domain-specific preprocessing (such as DICOM windowing, histogram equalization) is applied to maintain consistency with teacher model pipelines.

\subsubsection*{Downstream Benchmarks}

\begin{table}[htp]
\centering
\caption{Internal downstream benchmark statistics.}
\label{tab:supp_downstream}
\small
\setlength{\tabcolsep}{4pt}
\begin{tabular}{llrccl}
\toprule
\textbf{Modality} & \textbf{Dataset} & \textbf{\#Images} & \textbf{\#Classes} & \textbf{Metric} & \textbf{Split} \\
\midrule
\multirow{2}{*}{US} 
  & Thyroid   & 349  & 2 & AUC & 75:10:15 \\
  & Breast    & 780  & 2 & AUC & 75:10:15 \\
\midrule
\multirow{2}{*}{MRI}
  & ACL Tear  & 1,022  & 2 & AUC & 75:10:15 \\
  & Meniscus Tear & 4,201  & 2 & AUC & 75:10:15 \\
\midrule
\multirow{3}{*}{CT}
  & MediMeTA Axial    & 12,026 & 11 & ACC & Official \\
  & MediMeTA Coronal  & 12,026 & 11 & ACC & Official \\
  & MediMeTA Sagittal & 12,026 & 11 & ACC & Official \\
\midrule
Histo. & PCam & 327,680 & 2 & ACC & Official \\ 
\midrule
Derm. & HAM & 10,015 & 2 & ACC & Official \\ 
\bottomrule
\end{tabular}
\end{table}

We evaluate Med-RADIO on nine internal classification tasks spanning five imaging modalities (Tab.~\ref{tab:supp_downstream}). All datasets employ official train/test splits where provided; otherwise we adopt stratified 75:10:15 train/validation/test partitions following established medical imaging protocols~\cite{khattak2024unimed}. Evaluation metrics follow domain conventions: AUC for ultrasound/MRI; accuracy for CT, histopathology, and dermatology.

\subsection*{Teacher Model Selection Details}
We provide extended details regarding the selection criteria for both generalist and specialist teachers used in our multi-teacher distillation framework. We describe the factors considered in teacher inclusion—such as domain coverage, architectural compatibility, representational diversity, and empirical performance as well as additional ablation settings evaluating alternative teacher combinations.

\begin{table}[htbp]
\centering
\caption{Specialist-teacher candidates under linear probing. \textbf{Bold}: selected modality-specific specialist (UniMed-CLIP excluded from selection).}
\label{tab:teacher_selection}
\setlength{\tabcolsep}{2pt}
\begin{tabular}{l|l|cccccc|c}
\toprule
\multirow{2}{*}{Modality} & \multirow{2}{*}{Model} & \multicolumn{6}{c|}{Benchmarks} & \multirow{2}{*}{Avg} \\
\cmidrule{3-8}
                          &                        & \multicolumn{3}{c}{Thyroid (AUC)}    & \multicolumn{3}{c|}{Breast (AUC)}   &  \\
\midrule
\multirow{5}{*}{US}
  & USFM \cite{jiao2024usfm}                & \multicolumn{3}{c}{59.30} & \multicolumn{3}{c|}{64.73} & 62.01 \\
  & EchoCare \cite{zhang2025fully}            & \multicolumn{3}{c}{70.29} & \multicolumn{3}{c|}{77.94} & 74.12 \\
  & \textbf{URFM \cite{kang2025urfm}}       & \multicolumn{3}{c}{\textbf{74.15}} & \multicolumn{3}{c|}{\textbf{89.32}} & \textbf{81.74} \\
  & UniMed-CLIP (base) \cite{khattak2024unimed}  & \multicolumn{3}{c}{76.96} & \multicolumn{3}{c|}{79.34} & 78.15 \\
  & UniMed-CLIP (large) \cite{khattak2024unimed} & \multicolumn{3}{c}{79.18} & \multicolumn{3}{c|}{77.12} & 78.15 \\
\midrule
                          &                        & \multicolumn{3}{c}{ACL (AUC)}        & \multicolumn{3}{c|}{Meniscus (AUC)}   &  \\
\midrule
\multirow{4}{*}{MRI}
  & MR-CLIP \cite{avci2025mr}                        & \multicolumn{3}{c}{80.48} & \multicolumn{3}{c|}{87.33} & 83.91 \\
  & \textbf{Curia \cite{dancette2025curia}}                          & \multicolumn{3}{c}{\textbf{96.05}} & \multicolumn{3}{c|}{\textbf{92.69}} & \textbf{94.37} \\
  & UniMed-CLIP (base) \cite{khattak2024unimed}             & \multicolumn{3}{c}{97.30} & \multicolumn{3}{c|}{95.08} & 96.19 \\
  & UniMed-CLIP (large) \cite{khattak2024unimed}   & \multicolumn{3}{c}{97.05} & \multicolumn{3}{c|}{97.53} & 97.29 \\
\midrule
                          &             & \multicolumn{2}{c}{Ax. (ACC)}   & \multicolumn{2}{c}{Cor. (ACC)}  & \multicolumn{2}{c|}{Sag. (ACC)}  &  \\
\midrule
\multirow{3}{*}{CT}
  & \textbf{Curia \cite{dancette2025curia}} & \multicolumn{2}{c}{\textbf{76.70}}   & \multicolumn{2}{c}{\textbf{72.49}}  & \multicolumn{2}{c|}{\textbf{63.43}} & \textbf{70.87} \\
  & UniMed-CLIP (base) \cite{khattak2024unimed}  & \multicolumn{2}{c}{67.80}   & \multicolumn{2}{c}{59.22}  & \multicolumn{2}{c|}{62.78} & 63.27 \\
  & UniMed-CLIP (large) \cite{khattak2024unimed} & \multicolumn{2}{c}{76.38}   & \multicolumn{2}{c}{65.70}  & \multicolumn{2}{c|}{62.78} & 68.29 \\
\midrule
                          &             & \multicolumn{6}{c|}{PCam (ACC)}  &  \\
\midrule
\multirow{3}{*}{Histo.}
  & \textbf{GPFM \cite{ma2025generalizable}}  & \multicolumn{6}{c|}{\textbf{94.09}} & \textbf{94.09} \\
  & UniMed-CLIP (base) \cite{khattak2024unimed}  &  \multicolumn{6}{c|}{85.41} & 85.41 \\
  & UniMed-CLIP (large) \cite{khattak2024unimed} &  \multicolumn{6}{c|}{90.23} & 90.23 \\
\midrule
                          &             & \multicolumn{6}{c|}{HAM (ACC)}  &  \\
\midrule
\multirow{3}{*}{Derm.}
  & \textbf{PanDerm \cite{yan2025multimodal}}    & \multicolumn{6}{c|}{\textbf{89.19}} & \textbf{89.19} \\
  & UniMed-CLIP (base) \cite{khattak2024unimed}   &  \multicolumn{6}{c|}{86.26} & 86.26 \\
  & UniMed-CLIP (large) \cite{khattak2024unimed}  &  \multicolumn{6}{c|}{91.15} & 91.15 \\
\bottomrule

\end{tabular}
\end{table}

Table~\ref{tab:teacher_selection} summarizes the results, with bold entries marking the selected modality-specific specialist (UniMed-CLIP rows are reported for reference but excluded from specialist selection). URFM ranks first among ultrasound specialists; Curia is the strongest dedicated specialist on MRI and CT; GPFM dominates histopathology; and PanDerm is selected for dermatology. Together with UniMed-CLIP (large) as the mandatory generalist anchor, these choices form the final teacher ensemble: UniMed-CLIP (large), URFM, Curia, GPFM, and PanDerm.

\subsection*{Teacher Feature Statistics Computation}

The statistics expose two opposing failure modes. On one end, URFM (ultrasound, $\mathrm{Disp}{=}0.003$) and PanDerm (dermatology, $0.127$) encode highly concentrated feature spaces: their raw angular losses are negligibly small relative to other streams and would be entirely suppressed in unweighted multi-teacher distillation. The dispersion denominator amplifies these signals by factors of up to $333\times$, ensuring the student receives meaningful ultrasound and dermatology gradients. On the other end, Curia on CT ($1.264$) and GPFM on histopathology ($0.807$) produce large raw angular losses due to high feature spread, which would otherwise dominate the joint objective and force all other teachers' signals to zero. Dividing by these large dispersion values proportionally attenuates their contribution, preventing CT-driven optimization collapse. 
This dual correction is empirically validated by the balanced KD loss ablation (Fig.~\ref{fig:three_ablations}(c)): without it, CT Coronal ACC drops by $-15.05$pp, and ultrasound Thyroid AUC collapses by $-14.39$pp. This pattern aligns with Curia CT dominance while URFM is suppressed.

\begin{table}[hb]
\centering
\caption{Offline angular dispersion $\mathrm{Disp}(\Theta)$ on $10{,}000$ samples per native stream (Eq.~\ref{eq:angular_dispersion}). \textbf{--}: not applicable.}
\label{tab:angular_disp}
\setlength{\tabcolsep}{9pt}
\begin{tabular}{l|ccccc}
\toprule
Modality & UniMed-CLIP & URFM & Curia & GPFM & PanDerm \\
\midrule
Mixed          & $0.673$ & \textbf{--} & \textbf{--} & \textbf{--} & \textbf{--} \\
Ultrasound     & $0.136$ & $0.003$     & \textbf{--} & \textbf{--} & \textbf{--} \\
MRI            & $0.209$ & \textbf{--} & \textbf{--} & \textbf{--} & \textbf{--} \\
CT             & $0.218$ & \textbf{--} & $1.264$     & \textbf{--} & \textbf{--} \\
Histopathology & $1.057$ & \textbf{--} & \textbf{--} & $0.807$     & \textbf{--} \\
Dermatology    & $0.173$ & \textbf{--} & \textbf{--} & \textbf{--} & $0.127$     \\
\bottomrule
\end{tabular}
\end{table}

\subsection*{Hyperparameters and Training Pseudocode}
\label{sec:supp_algorithm}

We provide comprehensive implementation specifications to ensure full reproducibility of Med-RADIO. Tab.~\ref{tab:supp_hyperparams} consolidates all hyperparameter settings across three training stages: pretraining distillation, linear probing evaluation, and full fine-tuning adaptation. Algorithm~\ref{alg:supp_distill} formalizes the multi-stream distillation procedure with step-by-step annotations, explicitly detailing the proportional batch assembly strategy, balanced angular loss computation, and gradient optimization workflow. Together, these specifications enable exact replication of our training pipeline and facilitate extension to alternative medical imaging domains or teacher ensemble configurations.

\begin{table}[htbp]
\centering
\caption{Hyperparameter settings for distillation, linear probing, and full fine-tuning.}
\label{tab:supp_hyperparams}
\small
\begin{tabular}{ll}
\toprule
\textbf{Parameter} & \textbf{Value} \\
\midrule
\multicolumn{2}{l}{\textit{Pretraining (Student Distillation)}} \\
~~Architecture & ViT-L/16 (304M params) \\
~~Input resolution & $224{\times}224$ \\
~~Global batch size & 192 (32 per teacher stream; 6$\times$A100 80GB, Accelerate fp16) \\
~~Optimizer & AdamW ($\beta_1{=}0.9$, $\beta_2{=}0.999$, WD$=0.05$) \\
~~Learning rate schedule & Cosine (peak $1{\times}10^{-4}$, warmup 1 epoch) \\
~~Gradient clipping & Max norm 1.0 \\
~~Total epochs & 10 \\
~~Translator head & MLP + LayerNorm (1024-d CLS $\to$ $d_T$) \\
~~Dispersion estimation & 10,000 samples per teacher \\
\midrule
\multicolumn{2}{l}{\textit{Downstream Linear Probing}} \\
~~Frozen encoder & ViT-L/16 (all layers frozen) \\
~~Classification head & Linear (CLS $\to$ \#classes) \\
~~Batch size / LR / Epochs & 256 / $1{\times}10^{-3}$ / 50 \\
~~Optimizer & AdamW (cosine schedule) \\
\midrule
\multicolumn{2}{l}{\textit{Downstream Full Fine-tuning}} \\
~~End-to-end training & Encoder + task head jointly optimized \\
~~Batch size / LR / Epochs & 64 / $5{\times}10^{-5}$ / 50 \\
~~Optimizer & AdamW (cosine schedule) \\
\bottomrule
\end{tabular}
\end{table}

\begin{algorithm}[htbp]
\small
\caption{Med-RADIO Multi-Teacher Distillation with Balanced Loss}
\label{alg:supp_distill}
\begin{algorithmic}[1]
\STATE \textbf{Input:} Student encoder $f_\theta$; frozen teachers $\{T_m\}_{m=1}^{6}$; translator heads $\{\phi_m\}_{m=1}^{6}$; angular dispersions $\{\mathrm{Disp}(\Theta_m)\}_{m=1}^{6}$; data streams $\{\mathcal{D}_m\}_{m=1}^{6}$
\STATE \textbf{Initialize:} $f_\theta$ from ImageNet-21k ViT-L/16; $\{\phi_m\}$ via Xavier uniform
\FOR{epoch $e = 1$ to $10$}
    \STATE Set learning rate $\eta_e$ via cosine schedule (peak $1{\times}10^{-4}$, warmup 1 epoch)
    \STATE Shuffle each stream $\mathcal{D}_m$ with seed $e$
    \FOR{iteration $t$}
        \STATE \textcolor{gray}{// --- Step 1: Assemble multi-stream batch ---}
        \FOR{each stream $m \in \{1,\ldots,6\}$}
            \STATE Compute proportional batch size: $B_m \gets \lfloor (|\mathcal{D}_m| / \sum_{m'} |\mathcal{D}_{m'}|) \cdot 192 \rfloor$
            \STATE Sample mini-batch $\mathcal{B}_m \sim \mathcal{D}_m$ of size $B_m$
        \ENDFOR
        \STATE Assign residual samples: $\mathcal{B}_{\text{mix}} \gets \mathcal{B}_{\text{mix}} \cup \text{Sample}(192 - \sum_m B_m)$
        \STATE Construct global batch: $\mathcal{B} \gets \bigcup_{m=1}^{6} \mathcal{B}_m$ (total 192 images)
        
        \STATE \textcolor{gray}{// --- Step 2: Forward pass through student and teachers ---}
        \STATE $\mathbf{z}_s \gets f_\theta(\mathcal{B})$ \quad \textcolor{gray}{// Student CLS tokens [192, 1024]}
        \FOR{each stream $m$}
            \STATE Extract modality-specific tokens: $\mathbf{z}_s^{(m)} \gets \mathbf{z}_s[\mathcal{B}_m]$ \quad \textcolor{gray}{// [B\_m, 1024]}
            \STATE Compute teacher features: $\mathbf{z}_{T_m} \gets T_m(\mathcal{B}_m)$ \quad \textcolor{gray}{// Frozen forward pass}
        \ENDFOR
        
        \STATE \textcolor{gray}{// --- Step 3: Translate and compute balanced angular loss ---}
        \STATE Initialize $\mathcal{L}_{\text{total}} \gets 0$
        \FOR{each stream $m$}
            \STATE Project to teacher space: $\hat{\mathbf{z}}_s^{(m)} \gets \phi_m(\mathbf{z}_s^{(m)})$ \quad \textcolor{gray}{// [B\_m, d\_T\_m]}
            \STATE Normalize: $\hat{\mathbf{z}}_s^{(m)} \gets \hat{\mathbf{z}}_s^{(m)} / \|\hat{\mathbf{z}}_s^{(m)}\|_2$, $\mathbf{z}_{T_m} \gets \mathbf{z}_{T_m} / \|\mathbf{z}_{T_m}\|_2$
            \STATE Compute pairwise angular distance: $\Theta^{(m)} \gets \arccos(\hat{\mathbf{z}}_s^{(m)} \cdot \mathbf{z}_{T_m}^\top)$ \quad \textcolor{gray}{// [B\_m, B\_m]}
            \STATE Aggregate: $\mathcal{L}_m \gets \frac{1}{B_m^2} \sum_{i,j} \frac{(\Theta^{(m)}_{i,j})^2}{\mathrm{Disp}(\Theta_m)}$ \quad \textcolor{gray}{// Eq.~\ref{eq:angular_loss}}
            \STATE $\mathcal{L}_{\text{total}} \gets \mathcal{L}_{\text{total}} + \mathcal{L}_m$
        \ENDFOR
        
        \STATE \textcolor{gray}{// --- Step 4: Backward pass and optimization ---}
        \STATE Compute gradients: $\nabla_\theta \mathcal{L}_{\text{total}}$, $\{\nabla_{\phi_m} \mathcal{L}_{\text{total}}\}$
        \STATE Clip gradients: $\text{clip}(\nabla_\theta, \text{max\_norm}=1.0)$
        \STATE Update parameters: $\theta \gets \theta - \eta_e \nabla_\theta \mathcal{L}_{\text{total}}$; $\phi_m \gets \phi_m - \eta_e \nabla_{\phi_m} \mathcal{L}_{\text{total}}$ for all $m$
    \ENDFOR
\ENDFOR
\STATE \textbf{Output:} Trained student encoder $f_\theta$ (discard translator heads $\{\phi_m\}$)
\end{algorithmic}
\end{algorithm}

\end{document}